\documentclass{article}
\usepackage{ijcai19}

\usepackage{times}
\usepackage{soul}
\usepackage{url}
\usepackage[hidelinks]{hyperref}
\usepackage[utf8]{inputenc}
\usepackage[small]{caption}
\usepackage{graphicx}
\usepackage{amsmath}
\usepackage{booktabs}
\usepackage{float}
\usepackage{pgf}
\usepackage{hyperref}
\usepackage{dblfloatfix}
\usepackage{subcaption}
\title{Obstacle Tower: A Generalization Challenge in Vision, Control, and Planning}

\author{
Arthur Juliani$^1$\footnote{Corresponding Author}\and
Ahmed Khalifa$^2$\and
Vincent-Pierre Berges$^1$\and
Jonathan Harper$^1$\and\\
Ervin Teng$^1$\and
Hunter Henry$^1$\and
Adam Crespi$^1$\and
Julian Togelius$^2$\And
Danny Lange$^1$\\
\affiliations
$^1$Unity Technologies\\
$^2$New York University\\
\emails
\{arthurj, vincentpierre, jharper, ervin, brandonh, adamc, dlange\}@unity3d.com,
\\
ahmed@akhalifa.com, 
julian@togelius.com
}

\begin{document}

\maketitle

\begin{abstract}
The rapid pace of recent research in AI has been driven in part by the presence of fast and challenging simulation environments. These environments often take the form of games; with tasks ranging from simple board games, to competitive video games. We propose a new benchmark - Obstacle Tower: a high fidelity, 3D, 3rd person, procedurally generated environment \footnote{\url{https://github.com/Unity-Technologies/obstacle-tower-env}}. An agent playing Obstacle Tower must learn to solve both low-level control and high-level planning problems in tandem while learning from pixels and a sparse reward signal. Unlike other benchmarks such as the Arcade Learning Environment, evaluation of agent performance in Obstacle Tower is based on an agent's ability to perform well on unseen instances of the environment. In this paper we outline the environment and provide a set of baseline results produced by current state-of-the-art Deep RL methods as well as human players. These algorithms fail to produce agents capable of performing near human level.
\end{abstract}
\section{Introduction}

\begin{figure*}[!t]
    \centering
    \includegraphics[height=110px]{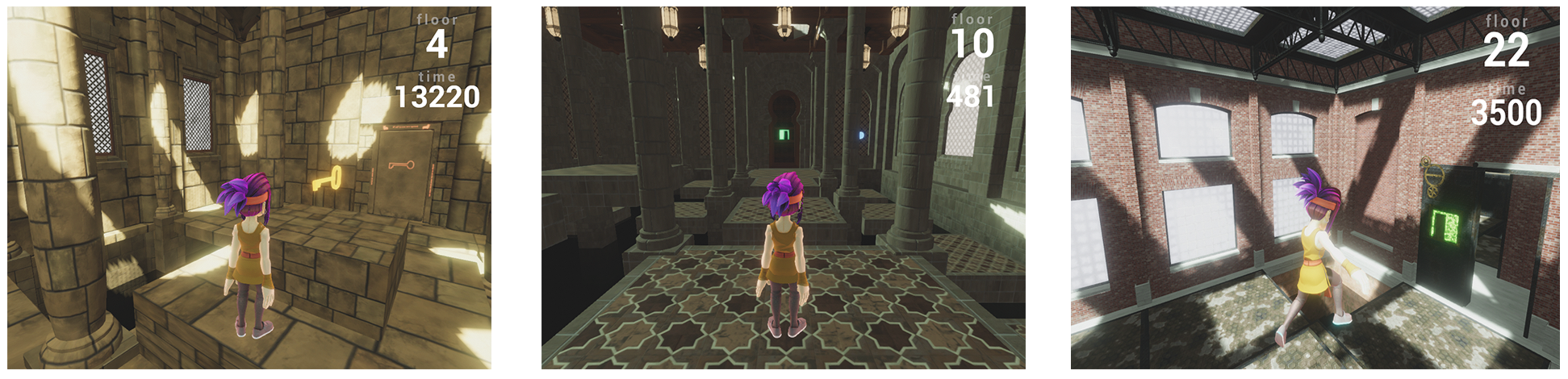}
    \caption{Examples of agent observations in the Obstacle Tower at different floor levels. [Left] Early floor is rendered in the \emph{Ancient} theme. [Middle] Intermediate floor is rendered using the \emph{Moorish} theme. [Right] Later floor is rendered in \emph{Industrial} theme.}
    \label{fig:visualExampleAgent}
\end{figure*}

\begin{figure*}[!t]
    \centering
    \includegraphics[height=110px]{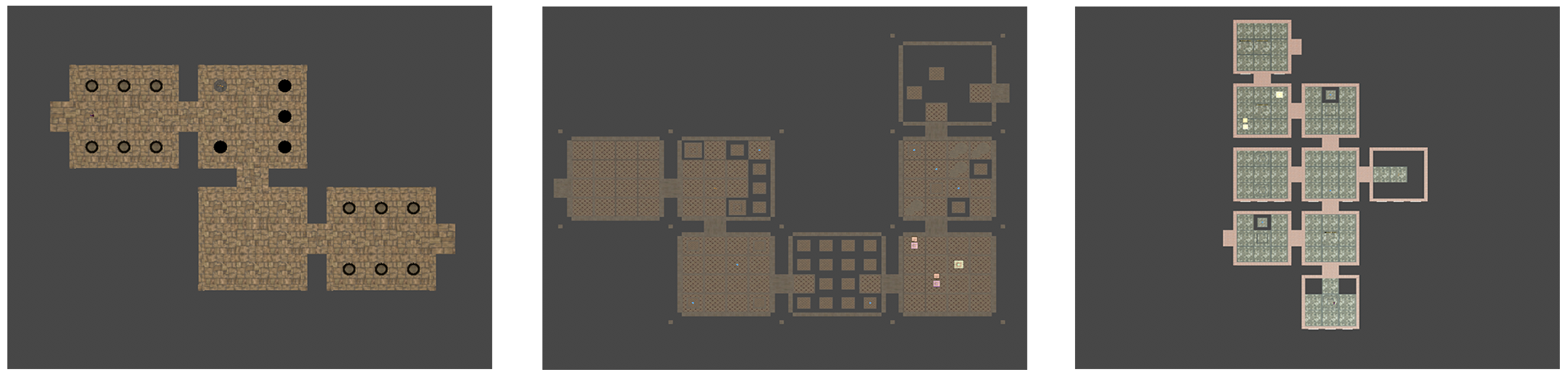}
    \caption{Examples of floor layouts in the Obstacle Tower at different floor levels. [Left] Early floor is rendered in the \emph{Ancient} theme. [Middle] Intermediate floor is rendered using the \emph{Moorish} theme. [Right] Later floor is rendered in \emph{Industrial} theme.}
    \label{fig:visualExampleFloors}
\end{figure*}

It is crucial for the development of artificial intelligence methods to have good benchmark functions, so that the performance of different methods can be fairly and easily compared. For tree search and reinforcement learning methods, the benchmarks of choice have often been based on games. Classic board games such as Checkers and Chess were prominent in AI research since its inception and spurred the development of many important techniques; for example, the first reinforcement learning algorithm was developed to play
Checkers~\cite{samuel1959some}. 

In the last two decades, video games have increasingly been used as AI benchmarks. In contrast to classic board games, video games require more frequent decision making, often in real-time settings, and additionally define more complex state spaces. They may or may not also have some combination of hidden information, stochasticity, complex interaction rules, and large branching factors. A number of benchmarks focus on classic 2D arcade games, as well as first-person shooters and racing games. These games all have limited branching factors, and the benchmarks built on them either make a low-dimensional processed observation of the environment available to the agent, or a fast forward model which allows for forward planning. This includes benchmarks based on Super Mario Bros~\cite{karakovskiy2012mario}, Minecraft~\cite{johnson2016malmo}, and Pac-Man~\cite{rohlfshagen2018pac}. The General Video Game AI competition is a special case of this, where agents are tasked with playing unseen 2D arcade-style games~\cite{perez2016general}.

A new generation of video game-based AI benchmarks do not provide agents with processed representations of the environment, but instead forces them to act based on the raw pixels, i.e. the screen output. The popularity of such benchmarks go hand-in-hand with the advent of reinforcement learning using deep neural networks as function approximators, so called deep reinforcement learning, as these deep networks are capable of processing high-dimensional input such as screen images. In particular, the Arcade Learning Environment (ALE), which is based on an emulation of the Atari 2600 video game console, became one of the more widely used reinforcement learning benchmark after it was demonstrated that Deep Q-learning could learn to play many of these games at a human-competitive level~\cite{bellemare2013arcade,mnih2015human}. 


The Atari 2600, on which ALE is based, is a very limited machine. It has 128 bytes of RAM, no video memory and games are typically 2 or 4 kilobytes of ROM; screen output is low-resolution 2D graphics. The lack of a system clock for seeding a pseudorandom number generator means that all games are deterministic. Having variability in the challenge, ideally through some kind of procedural content generation, is important for avoiding overfitting in reinforcement learning, and being able to evaluate what many AI researchers are actually interested in -  agent generalization~\cite{cobbe2018quantifying,zhang2018study,justesen2018illuminating}. Arguably, targeting generalization is necessary in order to make progress on artificial general intelligence, rather than just solving individual problems.

Recognizing these limitations, several game-based AI environments featuring raw pixel inputs have been proposed. The VizDoom competition and benchmark, based on the classic first-person shooter Doom is a prominent example~\cite{kempka2016vizdoom}. While it features a first-person perspective and complex gameplay, the age of the game means that the graphics are relatively primitive. Furthermore, the only kind of randomization is in enemy movement and item spawning, as the level topologies are fixed. Other recently introduced game-based AI benchmarks, such as the OpenAI Retro Challenge \cite{nichol2018gotta}, CoinRun \cite{cobbe2018quantifying}, and Pommerman \cite{resnick2018pommerman} all feature various kinds of environment randomization. They are however limited to providing 2D environment representation and only simple navigation challenges.

Obstacle Tower was developed specifically to overcome the limitations of previous game-based AI benchmarks, offering a broad and deep challenge, the solving of which would imply a major advancement in reinforcement learning. In brief, the features of Obstacle Tower are: 

\paragraph{High visual fidelity.} The environment is rendered in 3D using real-time lighting and shadows, along with much more detailed textures and model than previous benchmarks. See Figure \ref{fig:visualExampleAgent} for examples of the agents perspective.
\paragraph{Procedurally generated floors and rooms.} Navigating the game requires both dexterity and planning, and the floors within the environment are procedurally generated, making generalization a requirement to perform well during evaluation. See Figure \ref{fig:visualExampleFloors} for examples of floor layouts of various levels of the Obstacle Tower.
\paragraph{Physics-driven interactions.} The movement of the agent and other objects within the environment are controlled by a real-time 3D physics system.
\paragraph{Procedurally generated visuals.} There are multiple levels of variation in the environment, including the textures, lighting conditions, and object geometry. Therefore agents must be able to generalize their understanding of objects' appearance. 

\section{Obstacle Tower Environment}

Obstacle Tower provides recognizable and configurable observation spaces, action spaces, and reward functions. The environment itself relies heavily on procedural generation at multiple levels of interaction. To accommodate this, we propose a set of novel evaluation criteria specifically targeted at generalization, as well as outline the additional value provided by these design choices.

\subsection{Environment Specifications}

The Obstacle Tower environment uses the Unity platform and ML-Agents Toolkit \cite{juliani2018unity}. It can run on the Mac, Windows, and Linux platforms, and can be controlled via the OpenAI Gym interface for easy integration with existing DeepRL training frameworks \cite{brockman2016openai}.

\subsubsection{Episode Dynamics} The Obstacle Tower environment consists of up to 100 floors, with the agent starting on floor zero. All floors of the environment are treated as a single finite episode in the RL context. Each floor contains at the least a starting and ending room. Each room can contain a puzzle to solve, enemies to defeat, obstacles to evade, or a key to open a locked door. The layout of the floors and the contents of the rooms within each floor becomes more complex at higher floors in the Obstacle Tower, providing a natural curriculum for learning agents. Within an episode, it is only possible for the agent to go to higher floors of the environment, and not to return to lower floors. 

The episode terminates when the agent collides with a hazard such as a pit or enemy, when the timer runs out, or when the agent arrives at the top floor of the environment. The timer is set at the beginning of the episode, and completing floors as well as collecting time orbs increase the time left to the agent. In this way a successful agent must learn a behavior which is a trade off between collecting orbs and quickly completing floors of the tower in order to arrive at the higher floors before the timer ends.

\subsubsection{Observation Space} The observation space of the agent consists of two types of information. The first type of observation is a rendered pixel image of the environment from a third person perspective. This image is rendered in $168\times168$ RGB, and can be downscaled to $84\times84$. The second type of observation is a vector of auxiliary variables which describe relevant, non-visual information about the state of the environment. The elements which make up this auxiliary vector are: number of keys agent is in possession of, as well as the time left in the episode.

\subsubsection{Action Space} The action space of the agent is multi-discrete, meaning that it consists of a set of smaller discrete action spaces, of which the union corresponds to a single action in the environment. These subspaces are as follows: forward/backward/no-op movement, left/right/no-op movement, clockwise/counter-clockwise rotation of the camera/no-op, and no-op/jump. We also provide a version of the environment with this action space flattened into a single choice between one of 54 possible actions, whose size corresponds to the product of the sizes of all the sub-spaces in the multi-discrete case.

\subsubsection{Reward Function} Obstacle Tower has two reward function configurations: sparse and dense. In the sparse reward configuration, a positive reward of $+1$ is provided only upon the agent completing a floor of the tower. In the dense reward version a positive reward of $+0.1$ is provided for opening doors, solving puzzles, or picking up keys. In many cases even the dense reward version of the Obstacle Tower will likely resemble the sparsity seen in previously sparse rewarding benchmarks, such as Montezuma's Revenge \cite{bellemare2013arcade}. Given the sparse-reward nature of this task, we encourage researchers to develop novel intrinsic reward-based systems, such as curiosity, empowerment, or other signals to augment the external reward signal provided by the environment. 

\subsection{Procedural Generation of Floors}
Each floor of Obstacle Tower contains procedurally generated elements which impact multiple aspects of the agent's experience. These include lighting, textures, room layout, and floor plan. This proceduralism  ensures that for agents to do well on new instances of the Obstacle Tower, they must have learned general purpose representations of the task at the levels of vision, low-level control, and high-level planning. 

\subsubsection{Visual Appearance} On each floor of Obstacle Tower various aspects of the appearance of the environment are generated procedurally. This includes the selection of a visual theme which determines the textures and geometry to be used, as well as a set of generated lighting conditions. There are five distinct visual themes:  \emph{Ancient}, \emph{Moorish}, \emph{Industrial}, \emph{Modern}, and \emph{Future}. The lighting conditions include the direction, intensity, and color of the real-time light in the scene. 

\subsubsection{Floor layout} The floor layout is generated using a procedure inspired by Dormans~\cite{dormans2010adventures}. The floor layout generation is divided into two parts: a mission graph and a layout grid. 

The mission graph is responsible for the flow of the mission in the current level. For example: to finish the level the player needs to get a key then solve a puzzle then unlock the door to reach the stairs for the next level. Similar to Dormans, we used graph grammar which is a branch of generative grammar to generate the mission graph.

\begin{figure}
    \begin{subfigure}[b]{0.4\linewidth}
        \centering
        \includegraphics[height=80px]{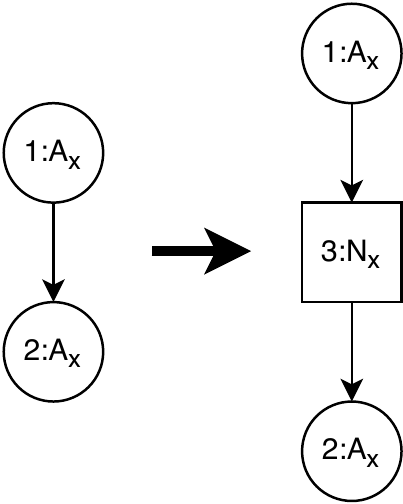}
        \caption{AddNormal Rule}
        \label{fig:addNormal}
    \end{subfigure}
    \begin{subfigure}[b]{0.6\linewidth}
        \centering
        \includegraphics[height=50px]{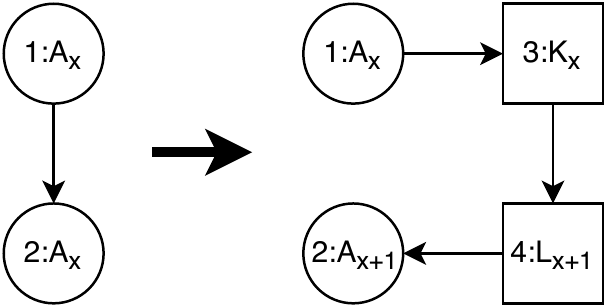}
        \caption{AddKeyLock Rule}
        \label{fig:addKeyLock}
    \end{subfigure}
    \begin{subfigure}[b]{0.45\linewidth}
        \centering
        \includegraphics[height=80px]{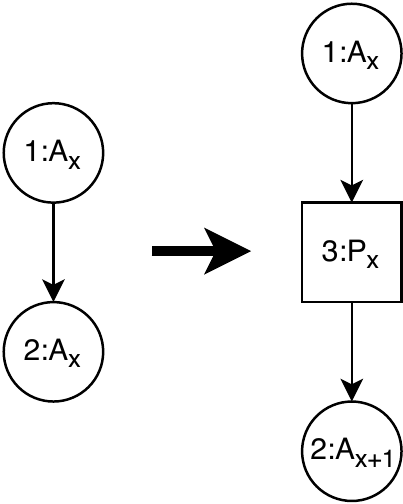}
        \caption{AddPuzzle Rule}
        \label{fig:addPuzzle}
    \end{subfigure}
    \begin{subfigure}[b]{0.45\linewidth}
        \centering
        \includegraphics[height=80px]{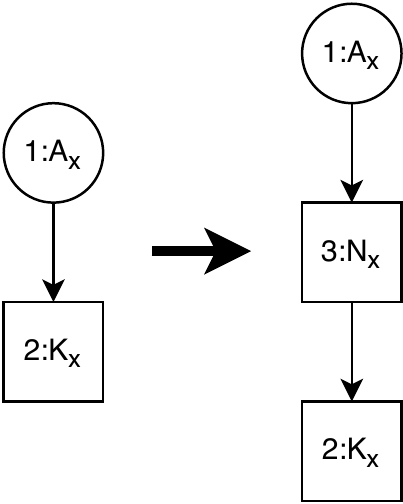}
        \caption{AddNormalKey Rule}
        \label{fig:addNormalKey}
    \end{subfigure}
    \caption{Four examples of Obstacle Tower mission graph rules.}
    \label{fig:otGrammarRules}
\end{figure}

Figure~\ref{fig:otGrammarRules} shows four of the graph rules used during generation. The letter in the node specifies the node type, while the small subscript number is the access level. The access level refers to how many locked doors must be opened in order to enter that room. The first number is used to make a mapping between nodes in the grammar. Circular nodes are considered as wild card nodes which means it can match any node. These rules are applied on the starting graph (consists of two connected nodes with access level of zero of type $start$ and $exit$) using what is called a graph recipe. A graph recipe is a sequence of graph grammar rules that are applied after each other to generate a level. The recipe contains some randomness by allowing each rule to be applied randomly more than once. In Obstacle Tower, the system uses different graph recipes for each group of levels to generate more complex floor layouts in later levels than the beginning levels.


After generating the mission graph, we then transform it into a 2D grid of rooms which is called layout grid. A simpler grammar called shape grammar~\cite{stiny1971shape} is used in the transformation process which is similar to Dorman's transformation process \cite{dormans2010adventures}. This layout grid is then directly used to generate the virtual scene which the agent navigates as a floor of the tower. See Figure \ref{fig:visualExampleFloors} for example images of the result of this process. 

\subsubsection{Room Layout} For the generation of the layout of each room within a floor, we used a template-based system similar to that used in the popular video game Spelunky. 
In this system each of the different room types, such as $Puzzle$ or $Key$ have their own set of templates from which specific room configurations are drawn from. In the case of the $Puzzle$ room, the agent must push a block from a starting location to a goal location, while avoiding intermediate obstacles. In the $Key$ rooms, there is a key somewhere in the room, along with potential obstacles. All templates consist of a grid of characters which represents the potential layout of the room. These grids can be either $3\times3$, $4\times4$, or $5\times5$. The specific placement of the modules and items within a room is based on these templates. The template can define the specific module or item to be placed in each position within the room, or define a category from which a specific module or item is drawn and placed probabilistically. In this way a finite number of templates can be used to generate a much larger number of possible room configurations. 

\subsection{Evaluation Criteria} 

It is essential that the evaluation of agent performance on environments such as the one described here be as reproducible and interpretable as possible. We provide three possible evaluation schemes. Because Obstacle Tower is designed to explicitly test the generalization ability of agents, we recommend evaluating using the latter two methods. This criteria described here is inspired by a recent set of recommendations by Henderson and colleagues \cite{henderson2017deep}. 

\paragraph{No Generalization.} It is possible to evaluate the performance of an agent on a single, fixed version of the Obstacle Tower. In this case we recommend explicitly reporting that the evaluation was performed on a fixed version of the Obstacle Tower, and also reporting performance on five random seeds of the dynamics of the agent. These seeds can be provided on environment reset, and condition the random number generator used to generate the tower definition.

\paragraph{Weak Generalization.} Agents should be trained on a fixed set of 100 seeds for the environment configurations. They should then be tested on a held-out set of five randomly selected tower configuration seeds not in the training set. Each should be evaluated five times using different random seeds for the dynamics of the agent (initial weights of the policy and/or value network(s)). 

\paragraph{Strong Generalization.} In addition to the requirements for weak generalization, agents should be tested on a held-out visual theme which is separate from the ones on which it was trained. In this paper we train on the \emph{Ancient} and \emph{Moorish} themes, and test on the \emph{Industrial} theme.

\subsection{Value as a Research Benchmark}
Obstacle Tower is designed to provide a meaningful challenge to current and future AI agents, specifically those trained using the pixels-to-control approach. There are four axes of challenge which we believe that this environment provides: vision, control, planning, and generalization. While various other environments and benchmarks have been used to provide difficult challenges for AI agents, this is to the best of our knowledge the first benchmark which combines all such axes of complexity. 
 

\subsubsection{Vision} The primary observation available to agents within the the Obstacle Tower is a rendered RGB image. Obstacle Tower contains high-fidelity real-time lighting, complex 3D shapes, and high-resolution textures. Furthermore, the floors in the environment are rendered in one of multiple different visual themes, such as \emph{Ancient} or \emph{Industrial}. These visual themes were chosen to provide a large amount of variation in the textures, colors, and 3D models that the agent would encounter. With the combination of high-fidelity visuals and increased visual variation, we expect models with much greater representational capacity than those used in A3C \cite{mnih2016asynchronous} or DQN \cite{mnih2015human} will be needed to perform well in the environment.

\paragraph{Generalization \& Vision.} Humans can easily understand that two different doors seen under different lighting conditions are still doors. We expect that general-purpose agents should have similar abilities, however this is not the case. In many cases agents trained under one set of visual conditions, and then tested on even a slightly different visual conditions perform much worse at the same task \cite{huang2017adversarial}. The procedural lighting and visual appearance of floors within the Obstacle Tower means that agents will need to be able to generalize to new visual appearances which they may never have directly experienced before. 

\subsubsection{Control} An agent in Obstacle Tower must be able to navigate through multiple rooms and floors. Each of these rooms can contain multiple possible obstacles, enemies, and moving platforms, all of which require fine-tuned control over the agent's movement. Floors of the environment can also contain puzzle rooms, which involve the physical manipulation of objects within the room in order to unlock doors to other rooms on the floor. While the action space of the agent is discrete, the environment itself uses continuous metrics for the position and velocity of objects, making the state space extremely large. We expect that in order for agents to perform well on these sub-tasks, the ability to model and predict the results of the agents actions within the environment will be of benefit.

\paragraph{Generalization \& Control.} The layout of the rooms on every floor are different on each instance of the Obstacle Tower, as such we expect methods which are designed to exploit determinism of the training environment, such as Brute \cite{machado2017revisiting} and Go-Explore \cite{clune2018} to perform poorly on the test set of environments. It is also the case that within a single instance of a Tower, there are elements of the environment which contain stochastic behavior, such as the movement of platforms and enemies. 

\subsubsection{Planning}
Depending on the difficulty of the floor, some floors of the Obstacle Tower require reasoning over multiple dependencies in order to arrive at the end room. For example, some rooms cannot be accessed without a key that can only be obtained in rooms sometimes very far from the door they open. In these cases, planning is required to ensure the agent takes the most efficient path between rooms.

\paragraph{Generalization \& Planning.} Due to the procedural generation of each floor layout within the Obstacle Tower, it is not possible to re-use a single high-level plan between floors. It is likewise not possible to re-use plans between environment instances, as the layout of each floor is determined by the environment's generation seed. Because of this, planning methods which require computationally expensive state discovery phases are likely not able to generalize to unseen floor layouts. 

\section{Preliminary Results}

In order to analyze the usefulness of Obstacle Tower benchmark, we conducted evaluations of the environment as well as agent and human performance within the environment. We evaluated human and agent performance within three distinct conditions, each designed to provide insight into the level of generalization ability that the human or agent possesses. We conducted this evaluation on version 1.0 of the Obstacle Tower which contains a maximum of 25 floors, and a limited subset of visual themes, floor configurations, and object types. We performed evaluation within the three conditions described under "Evaluation Criteria:" \emph{No Generalization} - training and testing on the same fixed environment, \emph{Weak Generalization} - training and testing on separate sets of environment seeds, and \emph{Strong Generalization} - training and testing on separate sets of environment seeds with separate visual themes.

\subsection{Environment Performance}

The ability to simulate at a high speed is important for ensuring that experimental iteration can take place at a reasonable pace. See Table \ref{table:environmentPerformance} for performance metrics detailing the average time it takes to perform an environment step from the Python interface. Note that this corresponds to five internal simulation steps, as the agent only requests decisions once every five simulation steps. This is similar to the concept of ``frame-skip" found in the ALE. These reported times also do not include model inference or training time typically involved in learning. These metrics were compared across floors and averaged over multiple seeds to provide a more general picture of performance. As expected, higher floors in the tower correspond to longer step times. This is due to the increasing complexity of the floor layouts on higher floors. Even on floor 20, we can still get roughly 350 simulation steps per second, with performance on the simplest floor around 500.

\begin{table}
\begin{center}
 \setlength\tabcolsep{5.5pt}
 \begin{tabular}{||c | c | c c c||} 
 \hline
 Floor & Per-Sec & Mean (Std) & Min & Max \\
 \hline\hline
 0 & 100.7 & 9.9 (1.6) & 9.6 & 10.1 \\ 
 \hline
 5 & 81.4 & 12.28 (1.6) & 11.5 & 12.9 \\ 
 \hline
 10 & 80.9 & 12.36 (1.7) & 11.5 & 14.3 \\ 
 \hline
 15 & 75.8 & 13.1 (4.2) & 12.2 & 14.4 \\ 
 \hline
 20 & 69.8 & 14.3 (5.2) & 13.3 & 15.7 \\ 
 \hline
\end{tabular}
\end{center}
\caption{Environment performance metrics on `n1-highmem-2' GCP instance with NVIDIA Tesla K80. Average steps-per-second and time-per-step (in ms). Averages recorded over five seeds each, and 500 steps per seed.}
\label{table:environmentPerformance}
\end{table}

\subsection{Human Performance}

In order to understand the expected quality of performance from a human-level agent, we conducted a series of evaluations with human play-testers. These were drawn from a pool of Unity Technologies employees who volunteered to participate in the evaluation process. These individuals did not have any particular background or established skill level with similar games. Overall fifteen participants took part in the evaluation. For human evaluation, training corresponds to the initial five minutes of playtime.

\begin{table}
\begin{center}
 \begin{tabular}{||c c c c||} 
 \hline
 Condition & Train & Test & Test (Max) \\ [0.5ex] 
 \hline\hline
 No Gen. & 15.2 (2.9) & 15.2 (2.9) & 22 \\ 
 \hline
 Weak Gen. & 12.3 (2.9) & 15.6 (3.5) & 21 \\
 \hline
 Strong Gen. & 12 (6.8) & 9.3 (3.1) & 20 \\
 \hline
\end{tabular}
\end{center}
\caption{Results of human evaluation on under different conditions. Performance results under \emph{Train} and \emph{Test} are reported as the \emph{mean (std)} of the number of floors solved in a single episode. Results reported under \emph{Test (Max)} correspond to maximum floor reached by a participant in each condition.}
\label{table:humanResults}
\end{table}

See Table \ref{table:humanResults} for human performance results. In the \emph{No Generalization} and \emph{Weak Generalization} conditions humans were able to solve an average of 15 floors during the test phase. Human participants performed slightly worse in the \emph{Strong Generalization condition}, however were still able to solve up to 20 floors in this condition as well, suggesting that humans are not able to perfectly transfer learned knowledge to new situations, but are also to do so with general success. As expected, these results show that humans are able to reuse knowledge gained during training to perform well on new unseen configurations of the environment. In fact we find that human performance in the \emph{Weak} and \emph{No} generalization conditions increases between the training and testing phases due to the ability of humans to continue to rapidly learn from small amounts of data. The additional difficulty of generalizing in the \emph{Strong generalization} condition is likely responsible for the decrease rather than increase between training and testing.

\subsection{Agent Performance}

We then turned to evaluation of agents trained using Deep RL. In particular we utilized the OpenAI Baseline implementation of Proximal Policy Optimization (PPO) \cite{schulman2017proximal,baselines}\footnote{\url{https://github.com/openai/baselines}} as well as the implementation of Rainbow provided by the  Dopamine library \cite{hessel2018rainbow,castro2018dopamine}\footnote{\url{https://github.com/google/dopamine}}. These two were chosen for being the standard implementations of current state of the art on-policy and off-policy algorithms. 

\begin{figure} [!t]
    \centering
    \scalebox{.4}{\input{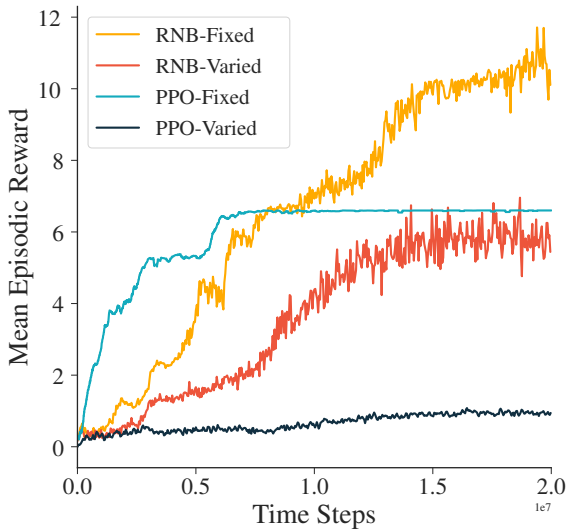}}
    \caption{Mean episodic reward received during training by agent trained using OpenAI Baseline PPO (PPO) and Dopamine Rainbow (RNB) in the Fixed and Varied training conditions.}
    \label{fig:agentResults}
\end{figure}

We performed training under the \emph{No Generalization} (Fixed) and \emph{Weak Generalization} (Varied) conditions, and performed evaluation within all three conditions. Based on initial experiments, we decided to conduct the evaluation using the "dense" reward function, due to a lack of significant learning with the "sparse" function. We utilized the default hyperparameters provided by each library for use with Atari benchmarks, in order to provide comparable results with evaluations performed on the ALE. We collected data in PPO using 50 concurrently running environments. In the case of Rainbow we collect data from a single environment running serially. We conducted training sessions spanning 20 million environment steps for PPO and Rainbow. 

See Figure \ref{fig:agentResults} for graphs of the mean reward during training of the two algorithms in both the varied and fixed conditions. We find that agents trained using either algorithm are able to solve fewer than 10 floors in both training conditions, with agents performing better in the fixed condition compared to the varied condition. This is to be expected, as the agents in the fixed condition are required to learn to solve only a single instance of Obstacle Tower, whereas agents in the varied condition are required to learn to solve a distribution of instances. In both fixed and varied training conditions we find that Rainbow outperforms PPO, suggesting that the Rainbow (Varied) agent has learned a more generalizable policy. We explicitly examine this generalizability in our evaluations.

\begin{table}
\begin{center}
 \setlength\tabcolsep{4.5pt}
 \begin{tabular}{||c c c c c||} 
 \hline
 Condition & PPO (F) & PPO (V) & RNB (F) & RNB (V) \\
 \hline\hline
 No Gen. & 5.0 (0.0) & 1.0 (0.0) & \textbf{7.0 (0.0)} & 4.8 (0.4) \\ 
 \hline
 Weak Gen. & 1.2 (0.4) & 0.8 (0.4) & 1.0 (0.7) & \textbf{3.4 (1.1)} \\
 \hline
 Strong Gen. & 0.6 (0.8) & 0.6 (0.5) & 0.6 (0.0) & \textbf{0.8 (0.8)} \\
 \hline
\end{tabular}
\end{center}
\caption{Results comparing trained models on three evaluation conditions. ``F'' corresponds to fixed training environment (one environment seed). ``V'' corresponds to varied training environment (100 environment seeds). Performance results are reported as the \emph{mean (std)} of the number of floors solved in a single episode.}
\label{table:agentEvaluation}
\end{table}

When the agents were benchmarked in the three evaluation conditions, we find that they consistently perform poorly compared to the human results, failing to reach even an average floor completion score of 10 in the fixed case, and floor 5 in the varied case. Interestingly, it is at floor 5 that the room mechanic of locked doors is introduced. We find that the varied agents are unable to solve this sub-task, and therefore are no longer able to make progress in the tower. This is possibly due to the lack of a long-term memory mechanism in these agents which would enable them to remember whether certain visited doors were locked or not. The partial success of the Rainbow (Fixed) agent in the \emph{No Generalization} condition is likely due to the ability to simply memorize the correct trajectory to the key and locked door locations. See Table \ref{table:agentEvaluation} for the full set of results.

The agents trained using Rainbow under the varied condition outperforms all other algorithms and training conditions in terms of evaluation performance on the weak and strong generalization conditions. One hypothesis for this is that the off-policy nature of the Rainbow algorithm allows for the agent to learn from a greater diversity of experiences, thus enabling better generalization to new conditions. 

As expected, agents in both training conditions perform significantly worse in the \emph{Strong Generalization} evaluation condition, with neither agent achieving an average floor completion rate of one. This result suggests that current Deep RL algorithms are very brittle with respect to their visual inputs.

\section{Discussion}

In this paper we have described the Obstacle Tower, a new research challenge for AI agents. Our preliminary results suggest that current state of the art methods achieve far less than human level performance on all experimental conditions. While the Rainbow agent is able to display limited generalization capabilities, they are significantly worse than those displayed by even the worst-performing human players. Furthermore, this difference was made apparent when using a dense reward function. Whether there exist methods capable of performing well using the sparse reward function in Obstacle Tower is an open question.

We believe that in order for learned agents to better perform on the task, fundamental improvements to the state of the art in the field will be required. We expect that these improvements will be more generally applicable beyond the Obstacle Tower itself, with impacting broader domains such as robotic navigation and planning.

The results presented in this paper correspond to version 1.0 of Obstacle Tower. We have since released a version 2.0 which included additional floors and configuration options for users. We plan to release a completely open source version of the Obstacle Tower project code in the coming months (version 3.0). This version will provide the ability to add additional state information such as a representation of the current floor layout, the freedom to modify the reward function, and the ability to add new module and item types into the procedural generation system. We hope that these extensions will allow the Obstacle Tower to not only be useful as a high-end benchmark of agents abilities, but also as a more general customizable environment for posing novel tasks to learning agents.

\subsection{Conclusion}

For the past few years the Arcade Learning Environment and other classic games have pushed the boundaries of AI research. We hope that the Obstacle Tower environment, with its focus on unsolved problems in vision, control, planning, and generalization, can serve the community in a similar way in the coming years. 

\section*{Acknowledgments}
The authors acknowledge the financial support from NSF grant (Award number 1717324 - "RI: Small: General Intelligence through Algorithm Invention and Selection."). 

We would additionally like to thank Leon Chen, Jeff Shih, Marwan Mattar, Vilmantas Balasevicius, and Yuan Gao for helpful feedback and support during the development and evaluation of this environment, as well as all the participants who took part in the human performance evaluation process.

\bibliographystyle{named}
\bibliography{ijcai19}

\end{document}